\documentclass[10pt,twocolumn]{article}
\usepackage[letterpaper,margin=0.75in]{geometry}
\usepackage[T1]{fontenc}
\usepackage{times}
\usepackage{graphicx}
\usepackage{microtype}
\usepackage{url}
\usepackage[numbers,sort&compress]{natbib}
\usepackage{amsmath}
\usepackage[table]{xcolor}  
\usepackage{booktabs}   

\usepackage{amssymb}
\usepackage{bbding}
\usepackage{multirow}
\usepackage{makecell}
\usepackage{array}
\usepackage{tcolorbox}
\usepackage{algorithm}
\usepackage{algorithmic}
\usepackage{caption}
\usepackage{comment}
\usepackage{environ}
\usepackage{listings}
\usepackage{placeins}
\usepackage{balance}
\tcbuselibrary{listings, breakable, skins}
\usepackage{hyperref}
\hypersetup{
  colorlinks=true,
  citecolor=blue,
  linkcolor=blue,
  urlcolor=blue
}
\graphicspath{{../}}

\makeatletter
\newcommand{\setcopyright}[1]{}
\newcommand{\copyrightyear}[1]{}
\newcommand{\acmYear}[1]{}
\newcommand{\acmDOI}[1]{}
\newcommand{\acmConference}[4][]{}
\newcommand{\acmISBN}[1]{}
\newcommand{\acmSubmissionID}[1]{}
\newcommand{\ccsdesc}[2][]{}
\excludecomment{CCSXML}
\newcommand{\@authorblocks}{%
  \resizebox{0.98\textwidth}{!}{%
    \mbox{\large
    Jiali Wen\textsuperscript{2,*}\quad
    Hongxia Gao\textsuperscript{1,3,4,*,\textdagger}\quad
    Litao Li\textsuperscript{2,*}\quad
    Yixin Chen\textsuperscript{2}\quad
    Kaijie Zhang\textsuperscript{2}\quad
    Qianyun Liu\textsuperscript{2}\quad
    Xiaoqin Wen\textsuperscript{2}}%
  }\par
  \vspace{0.65em}
  \footnotesize
  \textsuperscript{1}Xi'an Jiaotong University\quad
  \textsuperscript{2}South China University of Technology\quad
  \textsuperscript{3}Shenzhen Loop Area Institute\quad
  \textsuperscript{4}Pazhou Laboratory\par
  \vspace{0.45em}
  \textsuperscript{*}Equal contribution\quad
  \textsuperscript{\textdagger}Corresponding author: \href{mailto:hxgao@xjtu.edu.cn}{hxgao@xjtu.edu.cn}\par
  \vspace{0.25em}
  Student Project leaders: Jiali Wen (\href{mailto:1583412876@qq.com}{1583412876@qq.com}); Litao Li (\href{mailto:seonyee@foxmail.com}{seonyee@foxmail.com})\par
}
\newcommand{\email}[1]{}
\newcommand{\orcid}[1]{}
\newcommand{\authornote}[1]{}
\newcommand{\authornotemark}[1][]{}
\newcommand{\affiliation}[1]{}
\newcommand{\institution}[1]{}
\newcommand{\city}[1]{}
\newcommand{\state}[1]{}
\newcommand{\country}[1]{}
\newcommand{\shortauthors}{}
\newcommand{\@keywords}{}
\newcommand{\keywords}[1]{\gdef\@keywords{#1}}
\long\def\@storedabstract{}
\RenewEnviron{abstract}{\global\let\@storedabstract\BODY}
\renewcommand{\author}[1]{}
\renewcommand{\maketitle}{%
  \twocolumn[{%
    \begin{@twocolumnfalse}
    \begin{center}
      {\LARGE\bfseries \@title\par}
      \vspace{0.8em}
      {\normalsize\@authorblocks\par}
    \end{center}
    \vspace{0.6em}
    \end{@twocolumnfalse}
  }]%
  \noindent\textbf{Abstract}\par
  \noindent\@storedabstract\par
  \vspace{0.6em}
  \noindent\textbf{Keywords:} \@keywords\par
  \vspace{1.0em}
}
\makeatother

\newtcolorbox{promptbox}[1][]{%
  enhanced,
  colback=gray!4,
  colframe=black!55,
  fonttitle=\bfseries\small,
  title=#1,
  boxrule=0.45pt,
  arc=2pt,
  left=3pt,
  right=3pt,
  top=3pt,
  bottom=3pt,
  before skip=4pt,
  after skip=4pt
}

\lstdefinestyle{json}{
    basicstyle=\ttfamily\small,
    breaklines=true,
    frame=single,
    captionpos=b,
    numbers=none,
    backgroundcolor=\color{white},
    showstringspaces=false,
    literate={:}{{:}}{1}{,}{{,}}{1}
}
\AtBeginDocument{%
  }

\begin{document}

\title{OneFocus: Enabling Real-World X-ray Security Screening with a Unified Vision-Language Model}

\renewcommand{\shortauthors}{Wen et al.}

\begin{abstract}
X-ray contraband detection is critical for security in large-scale logistics and transportation, yet conventional detectors struggle to adapt to emerging contraband types and lack fundamental visual understanding. Vision-language models (VLMs) offer strong generalization but are hindered by the scarcity of high-quality X-ray image-caption data. To bridge this critical gap, we present \textbf{MMXray}, a meticulously curated benchmark of 52,124 image-caption pairs spanning 28 fine-grained classes of X-ray contraband. To enrich MMXray with realistic occlusion patterns, we further introduce \textbf{CleanDET}, a dedicated synthesis dataset containing clean foreground contraband images from 28 categories and background images with diverse density levels, together with \textbf{AnyContraSyn}, a controllable synthesis method designed to operate on CleanDET. We also develop \textbf{OnePipe}, an extensible pipeline for systematic data curation. Built on MMXray, we propose \textbf{OneFocus}, a unified VLM that supports four core tasks: visual question answering, contraband localization, classification, and image understanding. OneFocus achieves state-of-the-art performance in X-ray contraband understanding and demonstrates robust cross-domain generalization, establishing a strong vision-language baseline for security screening.
\end{abstract}

\ccsdesc[500]{General and reference~Evaluation}
\ccsdesc[300]{General and reference~Design}


\keywords{Multimodal Benchmark, Vision-Language Models, X-ray Security Screening, Visual Understanding.}


\maketitle

\section{Introduction}
X-ray contraband detection plays a critical role in high-stakes security environments such as airports, train stations, and subway systems. As passenger volumes continue to rise, regulatory requirements for prohibited item detection have become increasingly stringent. The growing diversity and volume of contraband items demand robust algorithms capable of comprehensive scene understanding under real-world screening conditions. Currently, most security checkpoints rely heavily on human operators to interpret X-ray images generated by screening systems or to validate outputs from basic detection pipelines, creating a bottleneck that limits scalability and consistency.

Existing approaches to X-ray contraband detection primarily rely on specialized object detection frameworks. Methods such as SDANet~\cite{pidray}, AENet~\cite{114Xray}, DOAM~\cite{opixray}, and LIM~\cite{hixray} enhance localization and classification through carefully designed feature fusion mechanisms. AO-DETR~\cite{aodetr} enriches semantic context in detection by pre-classifying object queries. Although effective for predefined categories, these methods exhibit limited generalization to unseen contraband types.

\begin{table}[h]
\centering
\caption{Comparison of open-source X-ray contraband detection datasets. \textbf{MultiModal} indicates the availability of image-caption pairs.}
\label{tab:datasets}
\small
\setlength{\tabcolsep}{8pt}
\renewcommand{\arraystretch}{1.1}
\resizebox{\linewidth}{!}{%
\begin{tabular}{lcccc}
\toprule
Dataset & Classes & Samples & MultiModal  \\
\midrule
GDXray~\cite{gdxray}  & 3 & 8,150 & \XSolidBrush  \\
SIXray~\cite{sixray}  & 6 & 8,929 & \XSolidBrush  \\
OPIXray~\cite{opixray}  & 5 & 8,885 & \XSolidBrush  \\
HiXray~\cite{hixray}  & 8 & 45,364 & \XSolidBrush  \\
CLCXray~\cite{clcxray} & 12 & 9,565 & \XSolidBrush  \\
PIXray~\cite{pixray}  & 15 & 5,046 & \XSolidBrush  \\
PIDray~\cite{pidray}  & 12 & 47,677 & \XSolidBrush  \\
114Xray~\cite{114Xray}  & 12 & \textbf{58,000} & \XSolidBrush  \\
DET-COMPASS~\cite{detcompass} & \textbf{370} & 1,928 & \XSolidBrush  \\ \hline
STCray~\cite{stcray} & 21 & 45,693 & \textcolor{red}{\checkmark}  \\
\rowcolor{gray!10}
\textbf{MMXray (Ours)} & 28 & 52,124 & \textcolor{red}{\checkmark}  \\
\bottomrule
\end{tabular}
}
\end{table}
Recent work has leveraged vision-language models for their open-vocabulary capabilities. OVXD~\cite{ovxd} adapts CLIP~\cite{clip} with a multimodal adapter to inject X-ray domain knowledge, enabling contraband classification and detection, but its high-level semantic abstraction limits localization accuracy. STING-BEE~\cite{stcray} fine-tunes LLaVA~\cite{llava15} using visual question answering data to improve contraband understanding, yet its reliance on multiple-choice questions and absence of open-ended answer supervision restrict comprehensive scene interpretation. These limitations highlight the need for richer, more structured supervision that jointly supports localization, reasoning, and open-ended language generation.

Although several public datasets exist for X-ray contraband detection~\cite{gdxray,sixray,opixray,hixray,clcxray,pixray,pidray,114Xray}, they universally lack image captions and multi-turn conversation data, resulting in a significant scarcity of multimodal data. While the STCray~\cite{stcray} introduced 45,693 captions for 21 contraband classes, its category coverage remains limited and the data is not sufficiently diverse. Furthermore, its captions are generated using fixed templates and simple descriptions, which leads to a lack of textual richness and may hinder the development of comprehension-based models.

To address these limitations, we introduce MMXray, a large-scale multimodal dataset for X-ray contraband scene understanding. MMXray comprises 52,124 image-caption pairs across 28 fine-grained contraband categories, carefully balanced to ensure uniform class distribution. The dataset is distinguished by four key properties: \textbf{Accuracy}, ensured through VLM-assisted caption generation followed by meticulous human verification; \textbf{Comprehensiveness}, as it supports diverse tasks including visual question answering, contraband localization, classification, and holistic image understanding, and enables rapid adaptation of captioners to new domains; \textbf{Diversity}, with images collected from real-world security settings such as airports, subway stations, parcel sorting lines, and augmented with physically plausible synthetic data; and \textbf{Strong}, providing sufficient data volume and category granularity to train robust models for X-ray interpretation.

\begin{figure}[h]
\centering
\includegraphics[width=\linewidth]{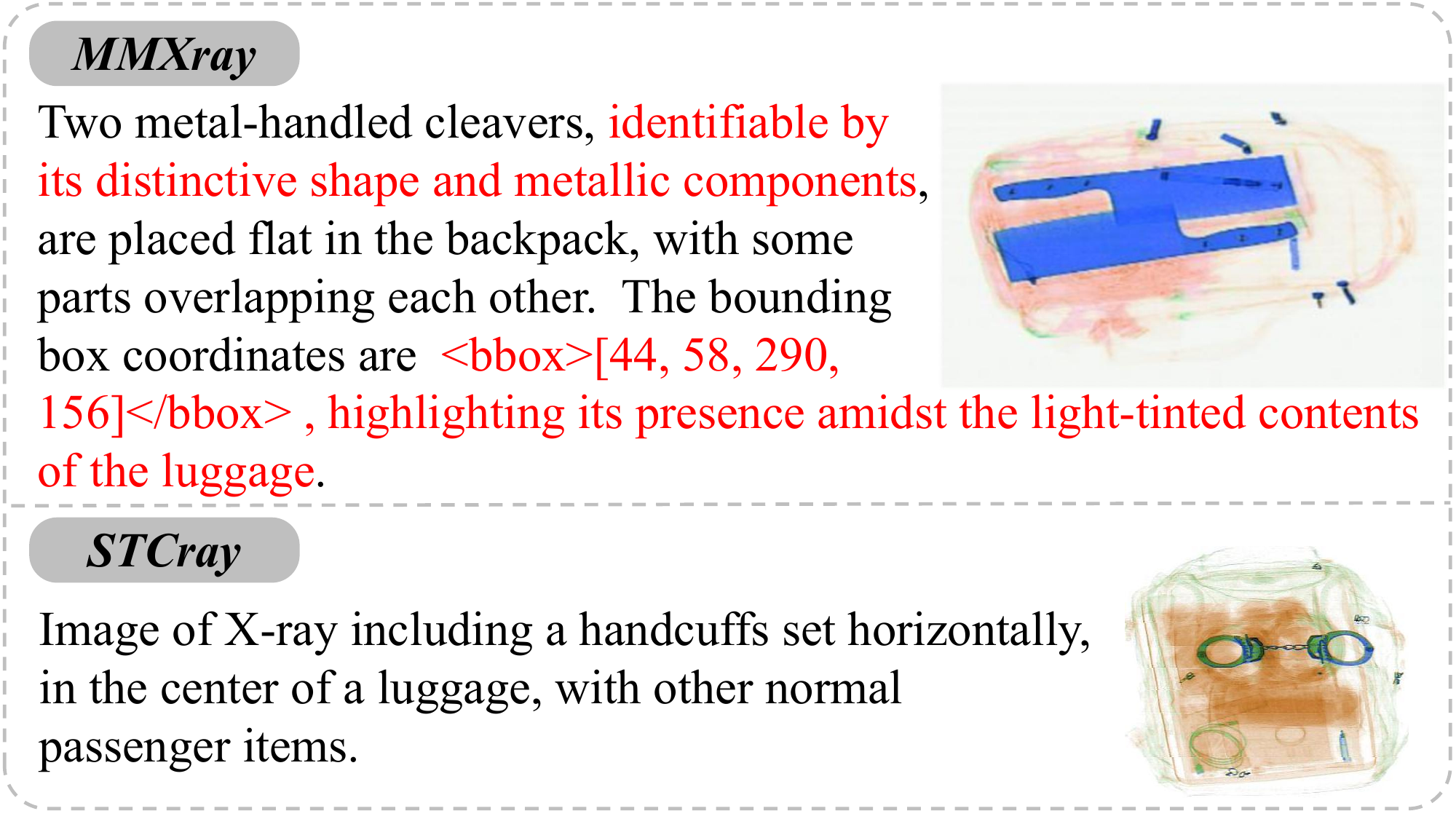}
\caption{MMXray is characterised by four features: accuracy, comprehensiveness, diversity, and Strong in terms of data diversity, data volume, image caption accuracy, and task coverage.}
\label{fig:mmxray}
\end{figure}

Building on MMXray, we propose OnePipe, a semi-automated captioning pipeline that leverages powerful vision-language models such as Qwen2.5-VL~\cite{Qwen2.5VL} to generate structured and diverse captions, refined through multi-stage human review and caption-specific model tuning. Finally, we train OneFocus on MMXray using a progressive learning strategy, endowing it with comprehensive scene understanding capabilities. OneFocus achieves state-of-the-art performance across all four tasks on multiple benchmarks, establishing a new standard for X-ray contraband understanding. In summary, our contributions are as follows:

\begin{itemize}
\item We introduce \textbf{MMXray}, a balanced multimodal X-ray dataset with 52,124 image-caption pairs across 28 contraband categories.
\item We propose AnyContraSyn, a controllable data synthesis method for supplementing occluded contraband images, and introduce CleanDET, a dedicated dataset that contains clean foreground images from 28 contraband categories and background images spanning diverse density levels for synthesis.
\item We propose \textbf{OnePipe}, a VLM-driven captioning pipeline for accurate and diverse X-ray contraband caption generation.
\item We propose \textbf{OneFocus}, a new multimodal baseline for X-ray contraband scene understanding that achieves state-of-the-art (SOTA) performance across multiple benchmarks in visual question answering, contraband localization, classification, and image understanding.
\end{itemize}

\section{Related Works}
\label{sec:rw}

\textbf{X-ray Contraband Datasets.} Early research used GDXray~\cite{gdxray} (industrial X-rays of baggage/welds/castings) for transfer learning. SIXray~\cite{sixray} offers >1M images across 6 prohibited classes, featuring severe class imbalance and occlusion. OPIXray~\cite{opixray} targets occluded scissors and proposes DOAM for de-occlusion. HiXray~\cite{hixray} expands to multi-category detection with LIM for background suppression. PIDray~\cite{pidray} provides pixel-level annotations for 12 contraband types, supporting detection, segmentation, and cross-domain evaluation. 114Xray~\cite{114Xray} includes 114K+ diverse scenes with and without contraband for realistic benchmarking. STCray~\cite{stcray} is the first multimodal dataset for conversational VLM training, offering structured textual descriptions of security screening scenes. DET-COMPASS~\cite{detcompass} provides the most contraband categories with 370 classes but only 1,928 images, lacks captions, and thus fails to support image understanding and unified multimodal tasks.

\textbf{Vision-Language Model.} Vision-language models (VLMs) have made substantial progress. CLIP~\cite{clip} enables open-vocabulary classification via contrastive learning on large-scale image–text pairs. BLIP-2~\cite{blip2} introduces a learnable Q-Former to improve cross-modal alignment efficiency and instruction-following capability. LLaVA~\cite{llava} establishes the dominant ViT + MLP + LLM paradigm, achieving strong emergent abilities with a lightweight projection layer and high-quality data. Qwen2.5-VL~\cite{Qwen2.5VL} extends reasoning to long-form videos and high-resolution images. InternVL3.5~\cite{internvl35} further enhances long-horizon reasoning through MPO~\cite{internvl35} and GSPO~\cite{gspo}.
However, existing VLMs are predominantly trained on natural RGB images and do not explicitly model key characteristics of X-ray imaging, including low texture, high penetration, material-dependent attenuation, and artifact-prone appearances. This limits their applicability to safety-critical contraband detection tasks.

\section{MMXray}

We introduce MMXray, a large-scale multimodal dataset for X-ray security inspection comprising 52,124 high-quality image-caption pairs across 28 fine-grained contraband categories. The dataset combines real-world X-ray scans from airports, subway stations, and parcel facilities with physically plausible synthetic images, covering a broad spectrum of prohibited items including knives, liquids, bullets, and handcuffs. MMXray captures a full range of occlusion conditions from minimal obstruction to severe overlap, enabling robust evaluation under realistic screening scenarios. Existing datasets often lack heavily occluded cases, which we address by synthesizing such images using a physics-aware method described in Section~\ref{sec31}. Captions are curated through expert annotation and an iterative refinement pipeline detailed in Section~\ref{sec32}. To enable stable acquisition of X-ray contraband scene knowledge and comprehensively enhance downstream vision tasks, we design six question-answer dimensions and construct hybrid-format question-answering data as outlined in Section~\ref{sec33}. To our knowledge, MMXray is the largest and most diverse multimodal dataset for X-ray security inspection, encompassing a wide variety of realistic operational conditions.

\begin{table*}[!tbp]
\centering
\caption{The 28 fine-grained categories of contraband in MMXray.}
\label{tab:categories}
\small
\begin{tabular}{*{4}{c}}
\toprule
Mobilephone & Powerbank & Handcuffs & Axe \\
Scissor & Baton & Bullet & Hammer \\
Gun & MetalhandleCleaver & PlastichandleCleaver & Pliers \\
Compressdgas & ColumnarblockBattery & PlasticbottleLiquid & GlassbottleLiquid \\
MetalbottleLiquid & UnfoldingKnife & FoldingKnife & StraightKnife \\
UtilityKnife & Multi-toolKnife & PlastichandleKnife & MetalhandleKnife \\
PlasticLighter & Laptop & Grenade & Fireworks \\
\bottomrule
\end{tabular}
\end{table*}

\subsection{Synthetic Data Acquisition: AnyContraSyn}
\label{sec31}

X-ray imaging fundamentally differs from natural-image formation. While data augmentation techniques like Mixup are commonly applied to natural images, they are ill-suited for X-ray contraband scenarios, where physical fidelity is paramount. We find that directly applying Mixup~\cite{mixup} to X-ray images leads to unrealistic pixel attenuation and distorted material density responses.

Specifically, the pixel intensity \(P_{xy}\) at location \((x, y)\) in an X-ray image is governed by the incident X-ray intensity \(I_0\) and the material optical depth \(\phi=\mu x\), where $\mu$ is the linear attenuation coefficient and $x$ is the thickness, as described by the Beer-Lambert law~\cite{bl}:
\begin{equation}
P_{xy}=I_{0}\,e^{-\phi}.
\label{eq:pxy}
\end{equation}
Assuming an ideal X-ray source intensity \(I_0=255\) consistent with 8-bit image dynamics, a naive Mixup operation that linearly combines two contraband images with optical-depth maps \(\phi_1\) and \(\phi_2\) produces a pixel value of \(I_0(e^{-\phi_1}+e^{-\phi_2})\). This formulation incorrectly models material superposition. Physically, when materials overlap, the total optical depth is additive, yielding the correct intensity \(I_0 e^{-(\phi_1+\phi_2)}\).

\begin{figure}[h]
\centering

\begin{minipage}[b]{0.22\linewidth}
    \centering
    \includegraphics[width=\linewidth]{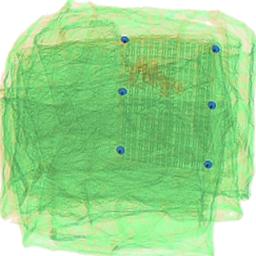}
    \\ 
    \includegraphics[width=\linewidth]{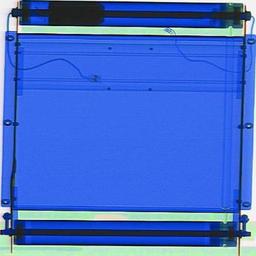}
    \\ 
    Background 
\end{minipage}
\hfill
\begin{minipage}[b]{0.22\linewidth}
    \centering
    \includegraphics[width=\linewidth]{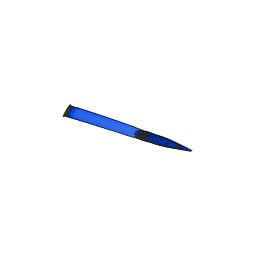}
    \\
    \includegraphics[width=\linewidth]{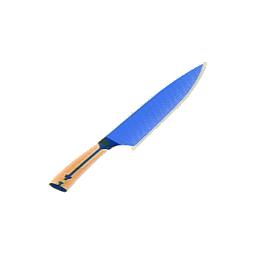}
    \\
    Contraband
\end{minipage}
\hfill
\begin{minipage}[b]{0.22\linewidth}
    \centering
    \includegraphics[width=\linewidth]{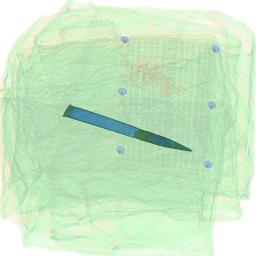}
    \\
    \includegraphics[width=\linewidth]{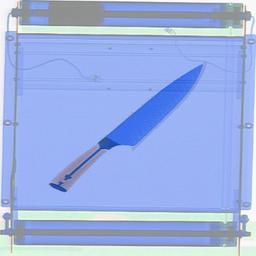}
    \\
    Mixup
\end{minipage}
\hfill
\begin{minipage}[b]{0.22\linewidth}
    \centering
    \includegraphics[width=\linewidth]{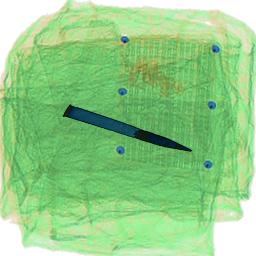}
    \\
    \includegraphics[width=\linewidth]{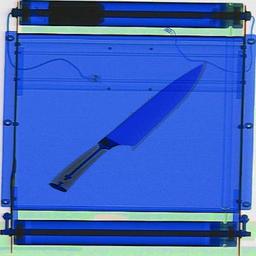}
    \\
    \textbf{Ours}
\end{minipage}

\caption{Synthetic data results. Background and Contraband as input. Our solution clearly outperforms Mixup in terms of photometric consistency and realism.}
\label{fig:mix}
\end{figure}

Standard Mixup~\cite{mixup} is physically inaccurate for X-ray imaging because it ignores the image formation process. To address this limitation, we propose \textbf{AnyContraSyn}, a physics-informed fusion method derived from the Beer-Lambert law for synthesizing physically plausible occluded contraband images. Given a foreground contraband image and a background image without contraband, AnyContraSyn first normalizes both images to $[0,1]$, then recovers their linear attenuation maps as
\begin{equation}
\phi_i = \ln \frac{P_i}{I_0},
\end{equation}
where $P_i$ denotes the observed X-ray intensity and $I_0$ is the incident intensity. Fusion is then performed additively in the attenuation domain as
\begin{equation}
\phi_{\mathrm{fused}} = \phi_{\mathrm{fg}} + \phi_{\mathrm{bg}},
\end{equation}
and the fused intensity image is reconstructed by
\begin{equation}
P_{\mathrm{fused}} = I_0 \exp \left(\phi_{\mathrm{fused}}\right).
\end{equation}
The final result is rescaled to $[0,255]$. Because fusion is conducted in the attenuation domain that underlies exponential X-ray formation, AnyContraSyn preserves realistic occlusion patterns, material interactions, and photometric consistency in non-occluded regions, thereby maintaining the original X-ray appearance of both the background and the contraband. This formulation also theoretically supports the composition of arbitrary contraband instances with arbitrary package backgrounds rather than only the single foreground-background case shown in the example.

AnyContraSyn performs best when the foreground image contains an isolated contraband instance with minimal occlusion and negligible background interference. However, such samples are rarely available in existing open-source X-ray datasets~\cite{114Xray,stcray,detcompass,pidray,opixray,hixray,gdxray}. To fill this gap, we collected single-contraband images and contraband-free background images from three real inspection scenarios, including airports, parcel security checkpoints, and subway stations. The background images span a wide density range and include metal baskets, plastic baskets, nylon parcels, wooden boxes, and plastic bags. We then used APSAM~\cite{apsam} to extract fine-grained contraband regions and remove redundant background content, resulting in \textbf{CleanDET}, a database containing clean contraband images from 28 categories and contraband-free package background images.

To further improve the controllability of AnyContraSyn, we introduce a mask-constrained placement strategy. Let $M_{\mathrm{fg}} \in \{0,1\}^{H_f \times W_f}$ and $M_{\mathrm{bg}} \in \{0,1\}^{H_b \times W_b}$ denote the foreground and background masks obtained from MobileSAMv2~\cite{MobileSAMv2}. We first anchor the contraband using $M_{\mathrm{fg}}$, then translate it to a candidate location $\mathbf{t} = (u,v)$ in the background mask. The translated foreground mask is denoted by
\begin{equation}
M_{\mathrm{fg}}^{\mathbf{t}}(x,y) = M_{\mathrm{fg}}(x-u,y-v).
\end{equation}
We define the valid overlap region as
\begin{equation}
M_{\mathrm{ov}}^{\mathbf{t}} = M_{\mathrm{fg}}^{\mathbf{t}} \odot M_{\mathrm{bg}},
\end{equation}
where $\odot$ denotes element-wise multiplication. A placement is considered valid only if the translated foreground mask is fully contained in the background support, namely
\begin{equation}
M_{\mathrm{fg}}^{\mathbf{t}} = M_{\mathrm{ov}}^{\mathbf{t}},
\end{equation}
or equivalently
\begin{equation}
\sum_{x,y} M_{\mathrm{fg}}^{\mathbf{t}}(x,y) = \sum_{x,y} M_{\mathrm{ov}}^{\mathbf{t}}(x,y).
\end{equation}
This constraint ensures that the contraband is entirely placed inside the package region, which better matches real deployment scenarios. We randomly sample valid locations that satisfy this condition and perform AnyContraSyn at those positions.

To align the foreground and background resolutions while preserving the original contraband scale and maximizing background purity, we embed the extracted foreground image into a white canvas with the same resolution as the background image. Formally, let $\Omega_{\mathrm{fg}}$ denote the support of the foreground image after translation. The padded foreground image $\tilde{P}_{\mathrm{fg}}$ is defined as
\begin{equation}
\tilde{P}_{\mathrm{fg}}(x,y) =
\begin{cases}
P_{\mathrm{fg}}(x-u,y-v), & (x,y) \in \Omega_{\mathrm{fg}}, \\
1, & \text{otherwise},
\end{cases}
\end{equation}
where the value $1$ corresponds to pure white in the normalized intensity domain. This operation stabilizes synthesis by matching image resolutions while avoiding the introduction of additional foreground clutter. Based on CleanDET, we generated 3,000 high-fidelity occluded contraband images with AnyContraSyn to augment the MMXray benchmark. The resulting synthetic images closely match real X-ray scans in both structural composition and intensity distribution.

\subsection{OnePipe}
\label{sec32}
The X-ray contraband domain suffers from a severe scarcity of high-quality image-caption paired data. Although the STCray~\cite{stcray} introduces a simple protocol that generates captions following the template \textit{``An X-ray image of [class] in [package] at [location]''}, such rigid templating, while covering basic attributes, produces semantically repetitive descriptions. This limits the diversity of vision-language model responses and increases the risk of overfitting.

\begin{figure*}[t]
\centering
\includegraphics[width=\linewidth]{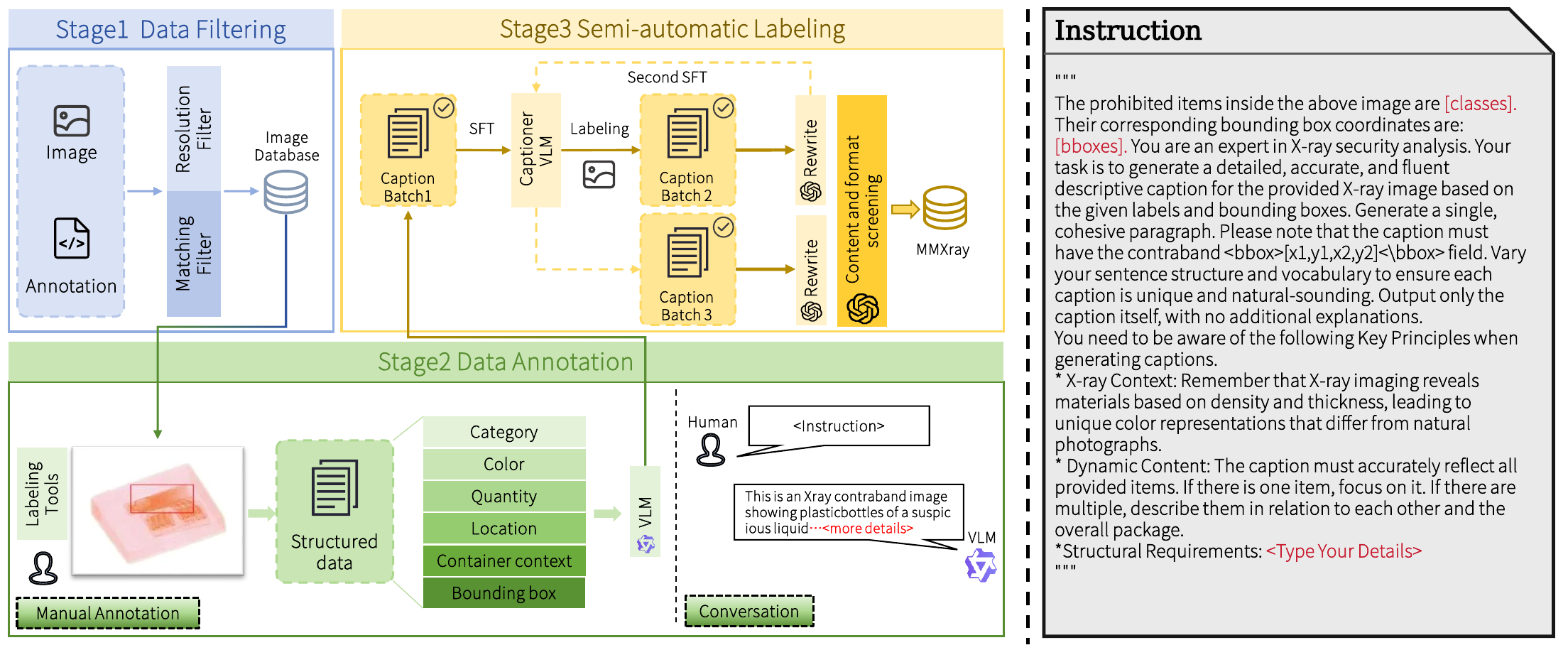}
\caption{OnePipe Workflow. Our pipeline comprises three stages: (1) Data Filtering, where images are screened by resolution, bounding boxes, and masks; (2) Data Annotation, in which domain-specific attributes of prohibited items are annotated via a custom tool and fed into Qwen2.5-VL~\cite{Qwen2.5VL} to generate structured captions; and (3) Semi-Automated labeling, where the Stage 2 image-caption pairs are used to fine-tune Qwen2.5-VL~\cite{Qwen2.5VL} into a dedicated captioner, whose outputs are then diversified via GPT-4o~\cite{gpt} rewrites before final refinement through continued training, enabling scalable, high-quality annotation of the full dataset.}
\label{fig:onepipe}
\end{figure*}

To address this limitation, we propose \textbf{OnePipe}, a scalable pipeline for acquiring, cleaning, and re-annotating X-ray contraband image-caption data. OnePipe is designed to generalize across diverse security screening scenarios. The full workflow is illustrated in Figure~\ref{fig:onepipe}.

Concretely, we first discard images with a height or width below 200 pixels. We further exclude ambiguous samples by ensuring consistency between the number of object bounding boxes and the bounding rectangles derived from segmentation masks, thereby preserving the model's counting capability. We then manually annotated 20,000 captions covering all contraband categories with a balanced class distribution. To enhance annotation efficiency and caption diversity, we focus on six key visual attributes: contraband category, color, quantity, approximate location, container context, and bounding box coordinates. Using Qwen2.5-VL-32B~\cite{Qwen2.5VL}, we design a specialized instruction prompt that incorporates fundamental X-ray imaging principles to guide the model in generating factually accurate and diverse captions while suppressing hallucinations.

This initial caption set is used to perform LoRA-based~\cite{lora} supervised fine-tuning on Qwen2.5-VL-7B~\cite{Qwen2.5VL}, yielding a dedicated Captioner model. The Captioner generates captions for an additional 10K images, which are then manually verified for attribute correctness. Verified captions are rewritten for linguistic diversity using the GPT-4o~\cite{gpt} API and subsequently added back to the training set for a second round of fine-tuning. The refined Captioner is then applied to annotate the remaining images, strictly following the same cleaning and re-annotation protocol.

\subsection{Hybrid-format Q\&A Data}
\label{sec33}

After obtaining image captions via OnePipe, we construct multi-turn conversations to support four core tasks: visual question answering (VQA), contraband location, classification, and image understanding. In contrast to STCray~\cite{stcray}, we employs a hybrid-format question-answering schema that includes both open-ended and multiple-choice questions, which better supports the model’s instruction-following ability and task-specific reasoning. We define six types that serve as the primary evaluation dimensions for assessing vision-language models in X-ray contraband scenarios: Instance Location, Instance Counting, Instance Identity, Instance Feature, Misleading, and Basic Understanding.

\textbf{Instance Location} leverages caption-derived candidate regions to query the approximate spatial position of contraband within a parcel. \textbf{Instance Counting} asks for the number of prohibited items, answered either as a numeric value or selected from discrete options. \textbf{Instance Identity} constructs classification-style questions by sampling from diverse contraband categories. \textbf{Instance Feature} prompts models to generate detailed open-ended responses describing object characteristics, encouraging comprehensive visual reasoning. \textbf{Misleading} introduces highly deceptive distractors that require strong discriminative ability to reject incorrect options. \textbf{Basic Understanding} evaluates fundamental domain knowledge by formulating questions grounded in X-ray imaging principles. We present example multiple-choice question-answer pairs for these six types in Table~\ref{tab:example_tasks}, with additional details provided in the supplementary material. Using GPT-4o~\cite{gpt}, each caption is expanded into 2 to 4 factually consistent conversation turns across these six dimensions. For open-ended answers, the prompting strategy explicitly encourages multi-step reasoning before producing the final response, ensuring rich and interpretable model outputs.
\begin{table}[h]
\centering
\footnotesize  
\setlength{\tabcolsep}{4pt}
\renewcommand{\arraystretch}{1.1}
\resizebox{\linewidth}{!}{%
\begin{tabular}{|>{\centering\arraybackslash}m{0.16\linewidth}|>{\raggedright\arraybackslash}m{0.30\linewidth}
|>{\centering\arraybackslash}p{0.54\linewidth}|}
\hline
\textbf{TaskType} & \textbf{Question} & \textbf{Options} \\
\hline
Instance Location & Is there a mobile phone located in this X-ray image? Where? &
\makecell[l]{A. Yes, it is on the upside of the image. \\
B. Yes, it is on the bottom of the image. \\
C. Yes, it is on the left of the image. \\
D. Yes, it is on the right of the image. \\
E. No mobile phone in the image.} \\
\hline
Instance Counting & How many batteries are there in the X-ray image? &
\makecell[l]{A. 0 \\
B. 1 \\
C. 2 \\
D. 3 \\
E. More than 3.} \\
\hline
Instance Identity & Which one is most likely to be the category of the X-ray scan image? &
\makecell[l]{A. Plasticbottleliquid. \\
B. Glassbottleliquid. \\
C. Powerbank. \\
D. Columnarblockbattery. \\
E. Grenade.} \\
\hline
Instance Feature & This is an X-ray scan containing powerbank. Is the feature of the powerbank a green rectangle? &
\makecell[l]{A. Yes. \\
B.  No.} \\
\hline
Misleading & Are there any laptops in this X-ray image?  &
\makecell[l]{A. Yes. \\
B.  No.} \\
\hline
Basic understanding & What is the most probable category for a sharp shape in an X-ray scan? &
\makecell[l]{A. Grenade. \\
B. Scissors. \\
C. Gun. \\
D. Fireworks. \\
E. Powerbank.} \\
\hline
\end{tabular}
}
\caption{Example tasks with questions and options. Only the single-choice format is shown here, see the Supplementary Material for open-ended responses.}
\label{tab:example_tasks}
\end{table}
For the MMXray test set, we retain the same six task categories as defined above. The test set contains only single-choice questions with two to five options, totaling 2,924 items. These questions are carefully curated. To increase task difficulty, we enhance the distractors by improving their semantic and visual similarity to the correct answer. This design targets the evaluation of basic visual-language understanding of X-ray contraband. To ensure full reproducibility of the MMXray data construction pipeline, we provide all prompts used in this stage as well as the principles for answer option recombination in the Supplementary Material.

\section{OneFocus}

We propose OneFocus, a versatile visual-language model capable of comprehensively addressing four key tasks in X-ray contraband scenarios: visual question answering, contraband localization, classification, and image understanding. OneFocus maintains strong competitiveness even in cross-domain settings.
\subsection{Model Architecture}

OneFocus largely reuses the architecture of Qwen2.5-VL~\cite{Qwen2.5VL}, consisting of a Vision Transformer~\cite{Qwen2.5VL} as the vision encoder, a two-layer MLP projector for semantic alignment, and the Qwen2.5 LM-7B~\cite{qwenlm} as the text decoder. The model is initialized from the pretrained weights of Qwen2.5-VL-7B. 

\subsection{Training Strategies}

\noindent\textbf{Training Preliminary Study.} Adapting general-purpose vision-language models (VLMs) to specialized domains such as X-ray security screening presents a fundamental challenge in balancing domain-specific adaptation with the preservation of general multimodal reasoning capabilities. Prior work has demonstrated that naive full fine-tuning often leads to catastrophic forgetting or inefficient use of computational resources, particularly when labeled data in the target domain is limited~\cite{domain,domain2}. In X-ray imaging contexts, where visual semantics diverge significantly from natural images, so we advocate for principled, component-wise adaptation strategies to achieve effective domain transfer while maintaining model robustness.

we conduct a systematic ablation study to identify the most effective trainable components for adapting Qwen2.5-VL-3B~\cite{Qwen2.5VL} to the X-ray contraband captioning task. Our experimental design isolates the contributions of the vision encoder (ViT), the large language model (LLM), and the vision-language projector. Following best practices in captioning evaluation, we sample 5,000 image–caption pairs from MMXray with uniform category distribution and deliberately exclude bounding box annotations to avoid introducing spatial bias into text generation metrics. and divide the training and validation sets in the ratio of $8:2$. We evaluate both full fine-tuning and parameter-efficient adaptation using Low-Rank Adaptation (LoRA)~\cite{lora} with rank 8 and alpha factor 32, the learning rate for full fine-tuning is $1 \times 10^{-5}$, and the learning rate for LoRA fine-tuning is $1 \times 10^{-4}$.

\begin{table}[t]
\centering
\small
\caption{Performance comparison of fine-tuning different modules. Here, \textbf{F} denotes full fine-tuning of the module, while \textbf{L} represents fine-tuning using LoRA with rank=8 and alpha=32. \textcolor{red}{Red} is the best result and \textcolor{blue}{Blue} is the second best result.}
\label{tab:finetune_comparison}
\begin{tabular}{l|ccc}
\hline
Module & SFT Strategy & BLEU-1 & ROUGE-1 \\
\hline
Base & - & 29.41 & 30.72 \\
Projector & F & 44.89 & 44.81 \\
Projector+ViT & F & 47.66 & 49.74 \\
Projector+LLM & F & 49.31 & 50.13 \\
ViT & F & 34.22 & 35.30 \\
LLM & F & \textbf{\textcolor{blue}{55.43}} & 54.61 \\
ViT & L & 54.86 & \textbf{\textcolor{blue}{56.14}} \\ \hline
\rowcolor{gray!10}
LLM & L & \textcolor{red}{\textbf{55.81}} & \textcolor{red}{\textbf{56.28}} \\
\hline
\end{tabular}
\end{table}

Table~\ref{tab:finetune_comparison} shows that fine-tuning only the language model yields the best performance. LoRA adaptation of the LLM alone achieves BLEU-1 and ROUGE-1 scores of \textbf{55.81} and \textbf{56.28}, matching or slightly exceeding full fine-tuning. The STING-BEE~\cite{stcray} strategy of jointly tuning the projector and the LLM performs worse than LLM-only fine-tuning, indicating that updating the projector provides no benefit and may even degrade optimization. Keeping the projector frozen leads to better alignment between visual features and textual outputs, suggesting that the pretrained projector already generalizes well to X-ray security imagery. Fine-tuning the ViT encoder offers only marginal gains over a fully frozen backbone, confirming that the vision model captures sufficient features for this domain. The main challenge lies not in visual representation but in mapping fixed visual features to domain-specific language concepts, a task most effectively addressed by adapting the language model alone.
\begin{table}[h]
\centering
\caption{Training Pipeline Configuration for Multi-Stage Fine-tuning.}
\label{tab:training_pipeline}
\resizebox{0.8\linewidth}{!}{%
\begin{tabular}{l | c c c}
\hline
\textbf{Categories} & \textbf{Stage 1} & \textbf{Stage 2} \\
\hline
Task & Captioning & Multi-Tasks  \\
Component & LLM & LLM  \\
Strategy & Full &LoRA \\
Tokens & $\sim$1M & $\sim$1.7M \\
Data & Reasoning Caption & Multi-Turns Q\&A \\
\hline
\end{tabular}
}
\end{table}

\textbf{Multi-stage fine-tuning: }We propose a two-stage fine-tuning strategy to adapt vision-language models to the X-ray contraband domain, as summarized in Table~\ref{tab:training_pipeline}. As shown in the first row of Table~\ref{tab:finetune_comparison}, VLMs exhibit poor zero-shot performance, indicating a fundamental lack of domain-specific understanding in the language model. To address this, Stage 1 performs full fine-tuning of the LLM on X-ray image captioning to establish a strong foundation of open-ended visual comprehension. In Stage 2, we inject LoRA into the adapted LLM and fine-tune it on multi-turn dialogues covering visual question answering, contraband localization, classification, and general image understanding, thereby preserving generative capability while enhancing task-specific accuracy.

\begin{table*}[t]
\centering
\caption{VQA performance across six subtasks. Average accuracy is reported as the overall metric. \textcolor{red}{Red} is the best result and \textcolor{blue}{Blue} is the second best result.}
\label{tab:vqa_results}
\resizebox{\textwidth}{!}{%
\begin{tabular}{l|ccccccc}
\toprule
Method & Location & Counting & Identity & Feature & Misleading & Basic Understanding & Avg\\
\midrule
LLaVA-1.5-7B~\cite{llava15} & 27.0 & \textcolor{blue}{\textbf{46.7}} & 37.0 & 36.7 & \textcolor{blue}{\textbf{68.9}} & 48.0 & 44.1\\
InternVL3.5-8B~\cite{internvl35} & \textcolor{blue}{\textbf{70.7}} & 43.9 & 28.5 & \textcolor{blue}{\textbf{89.3}} & 50.2 & \textcolor{blue}{\textbf{57.6}} & 56.7\\
Qwen2.5-VL-7B~\cite{Qwen2.5VL} & 63.2 & 45.2 & 33.1 & \textcolor{blue}{\textbf{89.3}} & 54.8 & 56.0 & \textcolor{blue}{\textbf{56.9}}\\
STING-BEE-7B~\cite{stcray} & 66.7 & 30.1 & \textcolor{blue}{\textbf{56.1}} & 36.5 & 50.0 & 47.2 & 47.8\\ \hline
\rowcolor{gray!10}
\textbf{OneFocus-7B (Ours)} & \textcolor{red}{\textbf{76.5}} &  \textcolor{red}{\textbf{60.7}} &  \textcolor{red}{\textbf{75.2}} &  \textcolor{red}{\textbf{94.8}} &  \textcolor{red}{\textbf{69.7}} & \textcolor{red}{\textbf{60.0}} & \textcolor{red}{\textbf{72.8}}\\
\bottomrule
\end{tabular}
}
\end{table*}

\section{Experiments}
\subsection{Implementation Details}

OneFocus is initialized from the pre-trained weights of Qwen2.5-VL-7B~\cite{Qwen2.5VL}. The entire training pipeline was implemented using the ms-swift~\cite{msswift} and conducted on a cluster of 8 NVIDIA A100 GPUs. The training process consists of two distinct stages. 

\textbf{Stage 1.} In this initial stage, we employed a batch size of 64. To stabilize early training and prevent potential gradient explosion, a cosine learning rate scheduler was utilized for the first 100 steps, with the final learning rate set to $2 \times 10^{-6}$. Memory efficiency was enhanced through the DeepSpeed-Zero2 optimization strategy. The model is trained for 1 epoch in this stage. \textbf{Stage 2.} For the subsequent fine-tuning stage, the batch size was increased to 128. We applied LoRA~\cite{lora} with a rank of 16 and an alpha value of 32 to all linear layers of the LLM, including those in the attention modules, providing sufficient adaptation capacity while maintaining parameter efficiency. To further optimize memory consumption for this more intensive stage, we leveraged DeepSpeed-Zero3. The model is trained for 2 epochs in this stage. \textbf{For Inference.} During inference, the temperature parameter for all models was set to 0.1 to ensure deterministic and accurate outputs. All experiments were repeated three times with random seeds 0, 1, and 42, and the results were averaged.

\subsection{Visual Question Answering}
\label{sec51}

For the VQA evaluation, we evaluate models across the six dimensions introduced in Sec~\ref{sec33}: Instance Location, Instance Counting, Instance Identity, Instance Feature, Misleading, and Basic Understanding. All questions are single-choice, with no open-ended responses. The evaluation uses the test sets from STCray and MMXray, containing a total of 40,589 question-answer pairs and 7,904 unique images. We assess VQA performance by computing the average accuracy across all subtasks. As shown in Table~\ref{tab:vqa_results}, OneFocus surpasses STING-BEE~\cite{stcray} on all metrics, achieving an absolute improvement of \textbf{25.0\%} in average accuracy.

\subsection{Classification}

We evaluate our method on three benchmark datasets for X-ray contraband classification: PIDray~\cite{pidray}, OPIXray~\cite{opixray}, and MMXray. PIDray contains 18,220 images spanning 12 contraband categories, covering a wide range of occlusion levels from minimal to severe, and includes both single- and multi-class detection scenarios. OPIXray comprises five fine-grained knife subcategories, enabling rigorous assessment of fine-grained visual discrimination. MMXray provides 3,200 high-quality images derived from real-world security screening environments, challenging models with complex multi-object and single-object classification tasks. We evaluate classification performance using the standard metrics F1 score and mean average precision (mAP). As shown in Table 6, OneFocus consistently outperforms prevailing vision-language models in contraband classification across multiple X-ray security benchmarks. Specifically, compared to the second best result, OneFocus achieves an absolute improvement of \textbf{+6.9\%} in F1 and \textbf{+8.0\%} in mAP on PIDray, \textbf{+7.0\%} in F1 and \textbf{+11.9\%} in mAP on OPIXray, and \textbf{+19.9\%} in F1 and \textbf{+13.7\%} in mAP on MMXray, highlighting its enhanced ability to understand X-ray contraband under real-world screening conditions.

\begin{figure*}[t]
\centering
\includegraphics[width=0.8\linewidth]{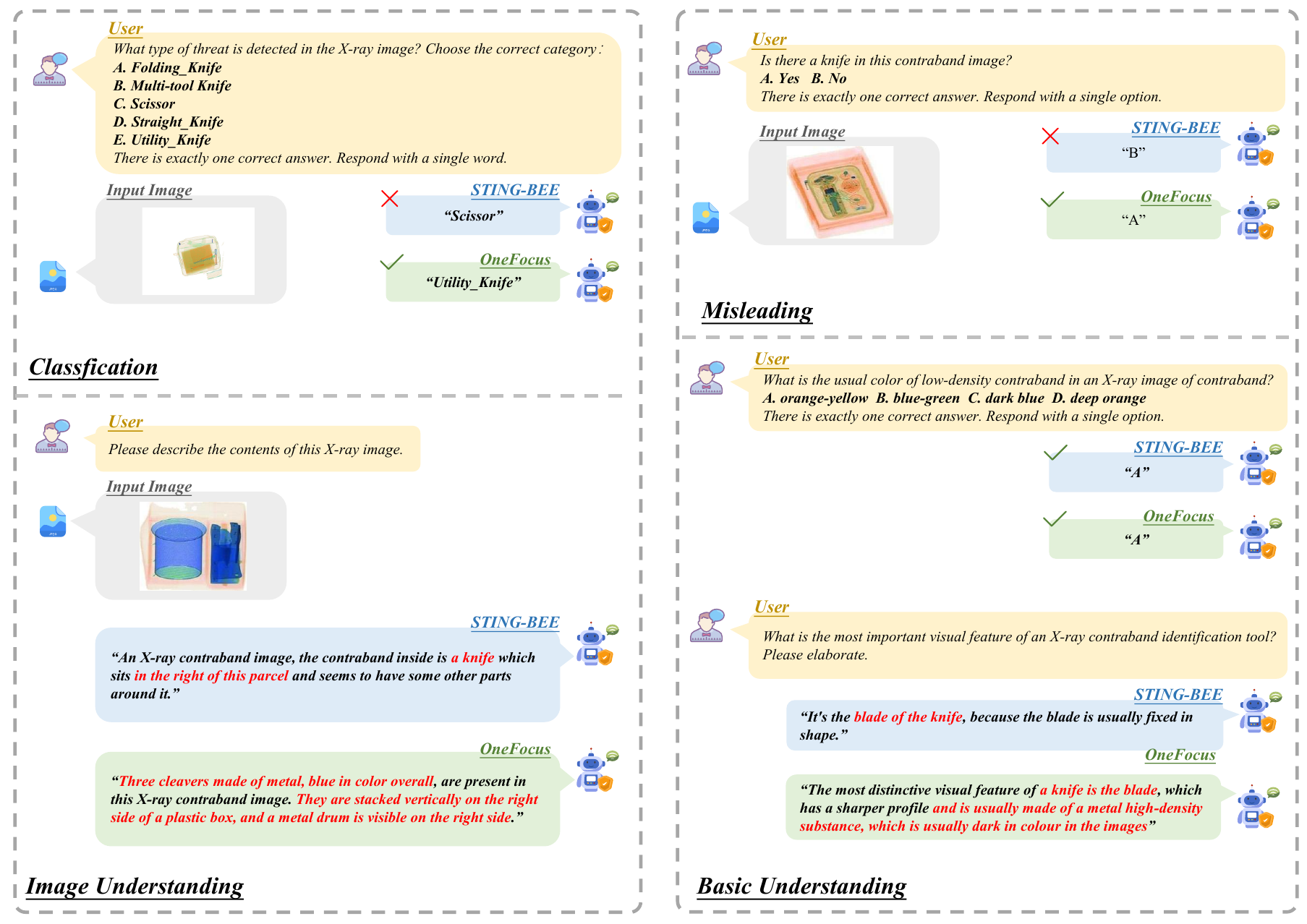}
\caption{Responses of OneFocus and STING-BEE~\cite{stcray} on Classification, Image understanding, Misleading, and Basic Understanding. OneFocus produces more stable and accurate responses. The \textcolor{red}{red} content represents important information for foundational understanding.}
\label{fig:answer}
\end{figure*}

\begin{table}[h]
\centering
\small
\caption{Classification performance (F1 and mAP, in \%) on PIDray~\cite{pidray}, OPIXray~\cite{opixray} and MMXray (Ours). \textcolor{red}{Red} is the best result and \textcolor{blue}{Blue} is the second best result.}
\label{tab:classification_results}
\resizebox{\linewidth}{!}{%
\begin{tabular}{@{}l|cccccc@{}}
\toprule
\multirow{2}{*}{Method} 
  & \multicolumn{2}{c}{PIDray}
  & \multicolumn{2}{c}{OPIXray} 
  & \multicolumn{2}{c}{MMXray} \\
\cmidrule(lr){2-3} \cmidrule(lr){4-5} \cmidrule(l){6-7}
  & F1 & mAP & F1 & mAP & F1 & mAP \\
\midrule
CLIP~\cite{clip}     & 2.1 & 3.9 & 2.5 & 5.4 & 10.3 & 11.0 \\
LLaVA-1.5~\cite{llava15}     & 12.5 & 21.1 & 10.3 & 11.1 & 38.9 & 48.2 \\
InternVL3.5~\cite{internvl35} & 18.1 & 20.6 & \textbf{\textcolor{blue}{22.4}} & \textbf{\textcolor{blue}{22.8}} & 26.7 & 28.0 \\
Qwen2.5-VL~\cite{Qwen2.5VL} & \textbf{\textcolor{blue}{22.6}} & \textbf{\textcolor{blue}{22.8}} & 20.8 & 17.4 & \textbf{\textcolor{blue}{46.1}} & \textbf{\textcolor{blue}{47.2}} \\ 
STING-BEE~\cite{stcray} & 8.7 & 18.7 & 11.9 & 13.1 & 35.3 & 45.6 \\ \hline
\rowcolor{gray!10}
\textbf{OneFocus (Ours) }                   & \textcolor{red}{\textbf{29.5}} & \textcolor{red}{\textbf{30.8}} & \textcolor{red}{\textbf{29.4}} & \textcolor{red}{\textbf{34.7}} & \textcolor{red}{\textbf{66.0}} & \textcolor{red}{\textbf{60.9}} \\
\bottomrule
\end{tabular}
}
\end{table}

\subsection{Contraband Localization}

\begin{figure}[h]
\centering
\includegraphics[width=\linewidth]{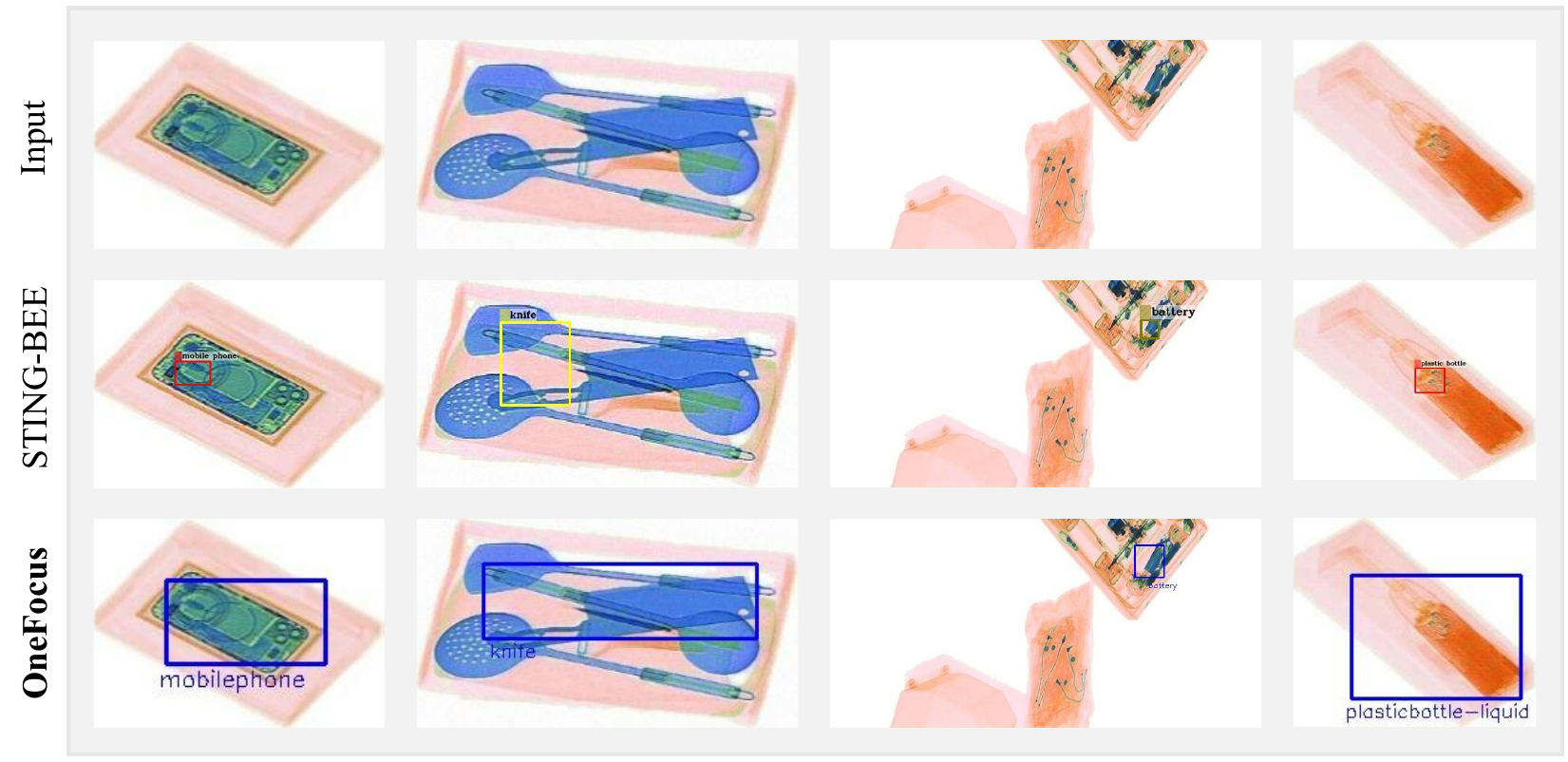}
\caption{Comparison of the localization ability of OneFocus and STING-BEE~\cite{stcray}, the first row is the input image, the second row is the localization result of STING-BEE, and the third row is the localization result of OneFocus, which is much more accurate in the localization of different scale categories.}
\label{fig:grounding}
\end{figure}

\begin{table}[h]
\centering
\footnotesize
\caption{Localization performance ($mAP_{50}$ and $mAP_{25}$) on PIDray~\cite{pidray} and MMXray (Ours). \textcolor{red}{Red} is the best result and \textcolor{blue}{Blue} is the second best result.}
\label{tab:detection_results}
\renewcommand{\arraystretch}{1.1}
\resizebox{\linewidth}{!}{%
\begin{tabular}{l|cccc}
\toprule
\multirow{2}{*}{Method}  & \multicolumn{2}{c}{PIDray} & \multicolumn{2}{c}{MMXray} \\
\cmidrule(lr){2-3} \cmidrule(lr){4-5}
 & $mAP_{50}$ & $mAP_{25}$ & $mAP_{50}$ & $mAP_{25}$ \\
\midrule
Faster R-CNN~\cite{faster} & 8.7 & 12.3 & 10.1 & 14.6 \\
YOLOv10~\cite{yolo} & 9.4 & 12.1 & 12.3 & 20.9 \\
DETR~\cite{detr} & 10.1 & 14.8 & 9.8 & 19.3 \\
GroundingDINO~\cite{groundingdino} & \textbf{\textcolor{blue}{17.5}} & \textbf{\textcolor{blue}{22.6}} & 19.6 & 24.5 \\
Qwen2.5-VL-7B~\cite{Qwen2.5VL} & 15.4 & 20.6 & 15.6 & 23.9 \\ 
STING-BEE-7B~\cite{stcray} & 12.8 & 19.7 & \textbf{\textcolor{blue}{20.5}} & \textbf{\textcolor{blue}{29.6}} \\ \hline
\rowcolor{gray!10}
\textbf{OneFocus-7B (Ours)} & \textcolor{red}{\textbf{18.5}}&\textcolor{red}{\textbf{31.2}} & \textcolor{red}{\textbf{32.2}} & \textcolor{red}{\textbf{51.6}} \\
\bottomrule
\end{tabular}
}
\end{table}
We evaluate open vocabulary object localization performance on the PIDray~\cite{pidray} and MMXray test sets, comparing OneFocus with a range of zero-shot detection methods. Both datasets encompass single- and multi-object detection scenarios and exhibit distinct color distributions due to differing X-ray imaging systems, providing a rigorous test of cross-domain generalization. We report mean Average Precision at IoU thresholds of 0.50 and 0.25 ($mAP_{50}$ and $mAP_{25}$), standard metrics for object detection. As shown in Table~\ref{tab:detection_results}, OneFocus achieves superior performance on both benchmarks. Compared to the best prior method, OneFocus achieves consistent improvements. On MMXray, it improves $mAP_{50}$ by \textbf{11.7\%} and $mAP_{25}$ by \textbf{22.0\%} compared to the second
best result. On PIDray~\cite{pidray}, it gains \textbf{1.0\%} in $mAP_{50}$ and \textbf{8.5\%} in $mAP_{25}$. Visualization results in Figure~\ref{fig:grounding} further demonstrate its superior capability in localizing diverse contraband.

\subsection{Image Understanding}
We evaluate image understanding capabilities of vision-language models on the test sets of STCray~\cite{stcray} and MMXray using BLEU-4 and ROUGE-L metrics. This choice is motivated by limitations of the single-choice VQA protocol adopted in~\cite{stcray}, which may overestimate true reasoning ability. Indeed, while STING-BEE achieves strong performance under that protocol (Sec~\ref{sec51}), its results drop significantly in our open-ended image understanding evaluation (Table~\ref{tab:image_understanding}), revealing a gap between single-choice accuracy and genuine scene comprehension.

\begin{table}[h]
\centering
\small
\caption{Image understanding performance on STCray~\cite{stcray} and MMXray test sets. \textcolor{red}{Red} is the best result and \textcolor{blue}{Blue} is the second best result.}
\label{tab:image_understanding}
\renewcommand{\arraystretch}{1.1}
\resizebox{0.8\linewidth}{!}{%
\begin{tabular}{l|cc}
\toprule
Method & BLEU-4 & ROUGE-L \\
\midrule
LLaVA-1.5-7B~\cite{llava15} & 1.5 & 18.1 \\
InternVL3.5-8B~\cite{internvl35} & 1.0 & 16.0 \\
Qwen2.5-VL-7B~\cite{Qwen2.5VL} & 1.0 & 17.2 \\
STING-BEE-7B~\cite{stcray} & \textbf{\textcolor{blue}{6.9}} & \textbf{\textcolor{blue}{27.7}} \\ \hline
\rowcolor{gray!10}
\textbf{OneFocus-7B (Ours)} & \textbf{\textcolor{red}{12.3}} & \textbf{\textcolor{red}{36.6}} \\
\bottomrule
\end{tabular}
}
\end{table}
In contrast, OneFocus demonstrates superior captioning quality, outperforming STING-BEE~\cite{stcray} by \textbf{5.4\%} in BLEU-4 and \textbf{8.9\%} in ROUGE-L. Qualitative examples in Figure~\ref{fig:answer} further illustrate that OneFocus generates accurate, detailed, and context-aware descriptions of X-ray scenes. This capability not only validates its strong image understanding but also enables direct use of OneFocus for high-quality image-caption pair generation, establishing a new foundation for multimodal data construction in X-ray security screening.

\subsection{Data Validity}
To decouple the contribution of MMXray from the inherent capacity of specific backbones, we conduct a controlled-variable analysis using both LLaVA-1.5 and Qwen2.5-VL (Table~\ref{re2}). While Qwen2.5-VL is pre-trained on a massive multi-modal corpus orders of magnitude larger than that of LLaVA-1.5, the vanilla Qwen2.5-VL is significantly outperformed by a LLaVA-1.5 model fine-tuned on MMXray. Specifically, LLaVA-1.5 trained on our dataset achieves a VQA accuracy of \textbf{55.6\%}, nearly closing the gap with the naive Qwen2.5-VL (56.9\%). More remarkably, localization performance (Loc) for LLaVA-1.5 surges \textbf{from 20.1\% to 36.3\%}. These results provide empirical evidence that the performance upper bound in specialized X-ray security tasks is primarily governed by the availability of high-quality, domain-specific instruction pairs rather than the sheer scale of the general-purpose pre-training corpus.

\begin{table}[h]
\centering
\caption{Cross-architecture validation on MMXray. \textbf{Loc} represents the mean of $mAP_{25}$ and $mAP_{50}$. Improvements demonstrate that MMXray significantly elevates the performance baseline, effectively bridging the gap between models pre-trained on different corpus scales.}
\label{re2}
\resizebox{\linewidth}{!}{
\begin{tabular}{lcccc}
\toprule
\textbf{Architecture} & \textbf{Data Strategy} & \textbf{VQA (\%)} & \textbf{Loc (\%)} & \textbf{$\Delta$ (VQA / Loc)} \\ \midrule
\multirow{2}{*}{LLaVA-1.5-7B~\cite{llava15}} & Naive & 44.1 & 20.1 & - \\
 & + MMXray & 55.6 & 36.3 & \textcolor{blue}{+11.5 / +16.2} \\ \midrule
\multirow{2}{*}{Qwen2.5-VL-7B~\cite{Qwen2.5VL}} & Naive & 56.9 & 19.8 & - \\
 & \textbf{OneFocus (Ours)} & \textbf{72.8} & \textbf{41.9} & \textcolor{blue}{+15.9 / +22.1} \\ \bottomrule
\end{tabular}
}
\end{table}

\subsection{Cross-domain Generalization}
The robustness of OneFocus is further evaluated through a cross-dataset zero-shot experiment on the STCray benchmark as presented in Table~\ref{re3}. When trained exclusively on MMXray with the strict exclusion of all STCray samples, OneFocus yields highly competitive results, achieving a VQA score of \textbf{48.0} and a Loc mAP of \textbf{43.1}. Notably, this performance surpasses established baselines even when those models are fine-tuned on related distributions. These findings indicate that MMXray enables the model to internalize fundamental X-ray imaging principles and contraband features. Consequently, the model demonstrates a superior ability to generalize to unseen data distributions and varying imaging hardware, rather than merely memorizing noise or biases inherent to a specific training set.

\begin{table}[h]
\centering
\footnotesize
\caption{Cross-dataset evaluation on STCray. OneFocus is trained \textbf{solely} on MMXray, strictly excluding all STCray-derived samples.}
\label{re3}
\resizebox{\linewidth}{!}{
\begin{tabular}{llcc}
\toprule
\textbf{Method} & \textbf{Training Set} & \textbf{VQA (Score)} & \textbf{Loc (mAP)} \\ 
\midrule
Qwen2.5-VL-7B~\cite{Qwen2.5VL} & Naive & 32.5 & 21.1 \\
STING-BEE~\cite{stcray} & STCray & 52.8 & 46.5 \\
STING-BEE~\cite{stcray} & MMXray & 42.4 & 38.8 \\
\midrule
\rowcolor{gray!10}
\textbf{OneFocus (Ours)} & \textbf{MMXray} & 48.0 & 43.1 \\
\bottomrule
\end{tabular}
}
\end{table}

\subsection{Ablation on Synthetic Data Usage}
\label{secsyn}
To investigate the contribution of the 3,000 synthetically augmented images in MMXray, we conduct a systematic ablation study across different training stages. These synthetic samples are specifically designed to simulate severe occlusion scenarios in security screening. As summarized in Table~\ref{tab:dataab}, incorporating synthetic data in both Stage 1 and Stage 2 yields the most significant performance gains, achieving an overall VQA accuracy of \textbf{72.8\%}. We observe that this strategy particularly enhances the model's ability to learn component-level features under color and contour distortions. These results validate the effectiveness of our synthetic augmentation in bridging the gap between ideal imaging and complex real-world occlusions.


\begin{table}[h]
\centering
\small
\caption{Ablation study on synthetic data usage across training stages. We report the average accuracy across all VQA tasks. \textbf{Bold} indicates the best performance.}
\label{tab:dataab}
\resizebox{0.6\linewidth}{!}{%
\begin{tabular}{cc|c}
\toprule
Stage 1 & Stage 2 & Avg. Acc. (\%) \\ \midrule
- & - & 64.7 \\
\checkmark & - & 65.3 \\
- & \checkmark & 70.1 \\
\rowcolor{gray!10}
\textbf{\checkmark} & \textbf{\checkmark} & \textbf{72.8} \\ \bottomrule
\end{tabular}
}
\end{table}

\section{Limitation}

\begin{figure}[h]
\centering
\includegraphics[width=\linewidth]{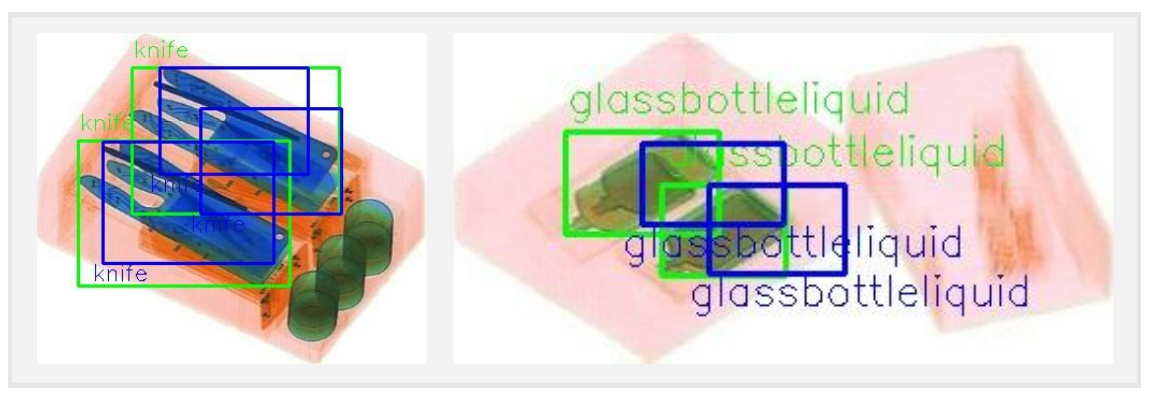}
\caption{Failure cases of OneFocus in contraband localization. The \textcolor{blue}{blue} bounding box is the prediction and the \textcolor{green}{green} bounding box is the ground truth.}
\label{fig:limit}
\end{figure}

Despite fine-tuning from a powerful vision-language foundation model, OneFocus still exhibits non-negligible localization errors when multiple contraband items share similar visual appearances, as shown in Figure~\ref{fig:limit}. This highlights the persistent challenge of adapting such models to precise spatial reasoning tasks.

These limitations stem from the fact that VLMs prioritize semantic alignment over spatial precision, yielding coarse visual representations and lacking coordinate awareness. As a result, they remain inherently weak at precise localization and often confuse overlapping contraband even after X-ray fine-tuning due to reliance on global context. OneFocus serves as a strong baseline for X-ray vision-language understanding. Despite its overall effectiveness, accurate contraband localization remains challenging, highlighting the need for future work on spatially aware VLM architectures.

\bibliographystyle{unsrtnat}
\bibliography{sample-base}


\newpage  
\appendix
\section*{Appendix}
Appendix contains the following: 
\begin{enumerate}
    \item Open-Source Plan (\ref{sec-com})
    \item Details of the manually annotated structured data (\ref{secb})
    \item Details of the construction of the training data for the multi-turn conversations (\ref{secc})
    \item Details of the VQA Evaluation (\ref{secd})
    \item Ablation on Synthetic Data Usage (\ref{app:secsyn})
\end{enumerate}

\section{Open-Source Plan}
\label{sec-com}
We commit to open-sourcing the full MMXray dataset, including image caption pairs, multi turn conversations, and evaluation benchmark, upon acceptance of this paper. A subset of image caption pairs and conversations is included in the Supplementary Material in JSON format for reference.

For OneFocus, we will release model weights and evaluation code in stages. Due to data security constraints, the training code will not be released.
We are actively expanding and improving MMXray with the aim of building a stronger multimodal benchmark for X-ray contraband detection. After the initial release, we will continue to add high quality multimodal data to MMXray and provide updated versions on a regular basis.

\section{Details of the manually annotated structured data}
\label{secb}
We built a manually annotation platform using Gradio and annotated X-ray images of contraband across six dimensions: Contraband category, Color, Quantity, Approximate location, Container context, and Bounding box coordinates. Below are the details for each dimension.
\begin{itemize}
    \item \textbf{Contraband category}: 28 classes defined in MMXray.
    \item \textbf{Color}: Ten colors reflecting material density under X-ray imaging, including light yellow, yellow, orange, light green, green, cyan, dark green, blue, dark blue, and black.
    \item \textbf{Quantity}: Integer counts from one to ten or coarse labels many and multiple. For heavily occluded cases where count is ambiguous, we randomly assign one of the two coarse labels.
    \item \textbf{Approximate location}: One of five spatial regions within the container: center, top, bottom, left, or right.
    \item \textbf{Container context}: Type of contrainer, such as suitcase, parcel, plastic box, or metal box.
    \item \textbf{Bounding box coordinates}: Loaded directly from COCO or YOLO format label files, visualized in the user interface, manually corrected when necessary, and automatically saved back to structured annotation files.
\end{itemize}

\begin{figure}[tbp]
\centering
\begin{promptbox}[Structured Data]
\footnotesize\ttfamily
\{ \\
  "category": "PlasticbottleLiquid", \\
  "color": "orange", \\
  "quantity": "2", \\
  "location": "center", \\
  "container context": "suitcase", \\
  "bbox": [[$x_{1}$, $y_{1}$, $x_{2}$, $y_{2}$], \\
  \hspace*{2em}[$x_{3}$, $y_{3}$, $x_{4}$, $y_{4}$]] \\
\}
\end{promptbox}

\caption{Examples of styles for manually annotated structured data.}
\label{fig:struct}
\end{figure}

An example of the fields in the final output json file is shown in Figure~\ref{fig:struct}.

\section{Details of the construction of the training data for the multi-turn conversations}
\label{secc}

In Stage 2 of training, we construct multi-turn conversations from image–caption pairs generated by OnePipe to align with the requirements of four visual tasks: Visual Question Answering, Classification, Contraband Localization, and Image Understanding. Our data combines both open-ended and closed-ended questions. This design addresses a key limitation observed in STING-BEE~\cite{stcray}, which relies solely on short VQA-style responses with templated user queries. Under such constraints, STING-BEE frequently fails to follow instructions or refuses to respond, producing invalid outputs such as \textit{“.”} in 10.9\% of cases on the MMXray benchmark. We find that training exclusively on short or closed-form conversations severely degrades a model’s instruction-following and reasoning capabilities.

\begin{figure*}[p]
\begin{promptbox}[Example Multi-turn Conversation]
\footnotesize
\centering
\includegraphics[width=0.34\linewidth]{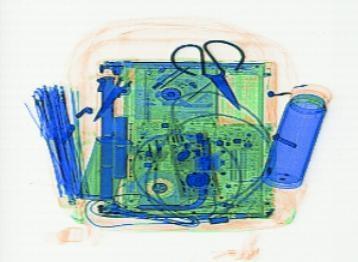}
\par\medskip
\raggedright
\noindent\textbf{MMXray Caption}:
\par
An X-ray scan image shows contraband including pressure, scissors, laptop, and metal bottle liquid. On the left side of the image, there is a partially occluded pressure vessel tank, with its surface material rendered in blue and internal gas in green, located at <bbox>[77,133,131,193]</bbox>. At the top of the image, there are two pairs of scissors. The scissors in the upper right side have a blue handle, indicating metallic meterial, and are located at <bbox>[169,37,250,112]</bbox>. The scissors on the left have a light-colored handle and are partially obscured by a high-density object, located at <bbox>[97,61,149,120]</bbox>. In the center of the image, there is a laptop with a blue-green color and circuit board texture, located at <bbox>[123,64,275,218]</bbox>. On the right side of the image, a metal bottle that may contain liquid is located at <bbox>[267,86,324,187]</bbox>. With other normal items together, all are placed in a single backpack.\par\medskip
\textbf{Multi-turn Conversations of Visual-Question Answering}:\par\smallskip

\textbf{User:} Classify the X-ray scan into exactly one or more categories.\\
\textbf{Assistant:} The image contains four categories: pressure, scissors, laptop, and metalbottleliquid.\par\smallskip
\textbf{User:} Can you tell me where is the laptop?\\
\textbf{Assistant:} Of course, the laptop is located at the center of the image.\par\smallskip
\textbf{User:} How many scissors are there in the X-ray image? \\
\textbf{Assistant:} There are 2 pairs of scissors visible in the image.\par\smallskip
\textbf{User:} What is the feature of the scissors? \\
\textbf{Assistant:} The tip of scissors is usually rendered blue, with a high density and a sharp edge.\par\smallskip
\textbf{User:} Are there any plastic bottles in the image? \\
\textbf{Assistant:} No,there is only metal bottle existing.Plastic bottle should be rendered orange-yellow,but the bottle in the image is colored blue.\par\smallskip
\textbf{User:}Are there any objects in the image that might be misjudged as belonging to other categories? \\
\textbf{Assistant:} Yes,the laptop in the image might be mistaken as a mobile phone because they have similar colors and circuit board textures.\par\smallskip
\textbf{User:}Are there any obstructed or hard-to-identify contraband in the image? \\
\textbf{Assistant:} Yes, there is a pressure vessel tank partially obscured in the image. In addition, the plastic handle of one pair of scissors is blocked by a high-density object.
\end{promptbox}
\caption{Sample Q\&A data generated by Qwen2.5-VL-32B~\cite{Qwen2.5VL}. VLM can derive accurate multi-turn conversations from the detailed reasoning captions provided in MMXray.}
\label{fig:multiturns}
\end{figure*}

To mitigate this, each caption is used to generate a multi-turn conversation containing both open-ended and closed-ended interactions, capped at four turns. For open-ended questions, we prompt a Qwen2.5-VL-32B~\cite{Qwen2.5VL} to generate queries and reasoning-rich answers along six dimensions: \textbf{Instance Location, Instance Counting, Instance Identity, Instance Feature, Misleading, and Basic Understanding}. For closed-ended questions, we provide 2 to 5 answer choices with exactly one correct option and enforce selection-only responses. The prompting strategy is illustrated in Figure~\ref{fig:prompt}. We recommend using vision-language models with at least 30B parameters and strong reasoning capabilities, as they demonstrate superior instruction adherence and support longer context windows, which are critical for generating high-quality multi-turn conversations. Multi-turn conversations generated through this pipeline is illustrated in Figure~\ref{fig:multiturns}, where only a subset of question answer pairs is shown for clarity. The full set exhibits substantially greater diversity and scale in practice.








\section{Details of the VQA Evaluation}
\label{secd}

For the VQA evaluation, we construct questions aligned with the same six task dimensions introduced in Section~\ref{secc}. However, a critical distinction must be emphasized: the data described in Section~\ref{secc} is intended solely for model training. It is coarse grained and includes both open ended and single choice questions. Moreover, the answer options in the training data are generated without deliberate design for discriminability or plausibility.

If the same data construction pipeline were applied to the evaluation benchmark, models could exploit superficial patterns or memorize option distributions, leading to overfitting that is difficult to detect through standard metrics. To ensure a rigorous and fair assessment of true visual language reasoning capability, we redesign all evaluation questions from scratch.

Specifically, every question in the VQA evaluation set is a single choice item, but the number of options varies per question. Neither the question phrasing nor the answer format follows a fixed template. Compared to the training data, we perform comprehensive restructuring of both question content and answer option design. This includes rephrasing queries to avoid lexical overlap with training prompts, introducing semantically similar distractors, and ensuring that correct answers cannot be inferred from option statistics alone. The goal is to elevate question difficulty and reduce the risk of shortcut learning, thereby providing a more reliable measure of model robustness and generalization in X-ray contraband understanding.

Following these principles, we curated a total of 2,924 evaluation samples. Each sample was manually verified to ensure semantic clarity, diverse phrasing, and high plausibility of distractors. Special attention was paid to minimizing lexical overlap with training prompts and preventing answer leakage through option patterns. This results in an evaluation set that prioritizes reasoning fidelity over surface statistics, enabling a more trustworthy assessment of model generalization in X-ray contraband understanding.

\subsection{Details of the question redesign}
In Instance Location, initial answers provide only coarse spatial cues such as “center,” “left,” or “right.” To improve localization precision, we augment the data with bounding box coordinates. The correct option is extracted directly from the structured annotations shown in Figure~\ref{fig:struct}, while distractors are synthesized by applying random perturbations to the ground truth box. Each perturbed box is constrained to maintain an intersection over union between 0.25 and 0.75 with the ground truth. This design encourages the model to develop fine grained spatial reasoning, and such coordinate based samples constitute 50\% of the Instance Location data. The whole process is shown in Algorithm~\ref{alg:instance_loc}.

\begin{algorithm}[tb]
\caption{Generate Options for Instance Location}
\label{alg:instance_loc}
\begin{algorithmic}[1]
\STATE \textbf{Input:} Ground truth box $b_{\text{gt}} = (x_1, y_1, x_2, y_2)$, IoU range $[\tau_{\min}, \tau_{\max}] = [0.25, 0.75]$
\STATE \textbf{Output:} Answer options $\mathcal{O} = \{b_{\text{gt}}\} \cup \{b_1, \dots, b_K\}$
\STATE $\mathcal{O} \gets \{b_{\text{gt}}\}$ \quad 
\FOR{$k = 1$ \textbf{to} $K$}
    \REPEAT
        \STATE Sample offsets $\Delta x_1, \Delta y_1, \Delta x_2, \Delta y_2 \sim \mathcal{U}(-\delta, \delta)$
        \STATE $b_{\text{cand}} \gets (x_1 + \Delta x_1,\ y_1 + \Delta y_1,\ x_2 + \Delta x_2,\ y_2 + \Delta y_2)$
        \STATE $\text{iou} \gets \text{IoU}(b_{\text{cand}}, b_{\text{gt}})$
    \UNTIL{$\tau_{\min} \leq \text{iou} \leq \tau_{\max}$}
    \STATE $\mathcal{O} \gets \mathcal{O} \cup \{b_{\text{cand}}\}$
\ENDFOR
\end{algorithmic}
\end{algorithm}

In Instance Counting, we generate questions that require exact enumeration of contraband items. We include both typical cases and challenging scenarios such as zero instances or densely packed objects. All answers are cross verified against the underlying instance annotations to ensure consistency and reliability. 

In Instance Identity. we supports two response formats: open-ended and single-choice. In the open-ended format, each answer includes a brief reasoning chain and the correct answer. In the single-choice format, we reorganize answer options after constructing the Hybrid-format Q\&A data. Specifically, the correct option remains unchanged. For distractors, we use GPT-4o~\cite{gpt} to rank candidate categories by textual semantic similarity to the correct answer category and select the top three as distractors to make single-choice questions more challenging. The specific instructions are shown in Figure~\ref{fig:redesign}.

Instance Feature evaluates a model's ability to recognize component level features of contraband. In each question the ground truth contraband category present in the image is explicitly provided. The model is then given a textual description of one or more component level attributes of the item in the X-ray image such as its color contour or other structural characteristics. This description may be factually correct or deliberately inaccurate. The model must choose between Yes and No to indicate whether the description matches the visual evidence.
\begin{table*}[t]
\centering
\caption{Model accuracy under varying proportions of ``No'' answer questions in Misleading. 
We vary the ground-truth ``No'' ratio from 0\% (mimicking natural imbalance) to 100\% (extreme bias scenario). Where \textbf{Naive} represents the benchmark native scale provided by MMXray. \textcolor{red}{Red} indicates the best result. }
\label{tab:no_ratio_ablation}
\resizebox{\textwidth}{!}{%
\begin{tabular}{l|cccc}
\toprule
 ``No'' GT & LLaVA-1.5-7B~\cite{llava15} & Qwen2.5-VL-7B~\cite{Qwen2.5VL} & InternVL3.5-8B~\cite{internvl35} & \textbf{OneFocus (Ours)} \\
\midrule
Naive & 68.9 & 54.8 & 50.2 & \textbf{\textcolor{red}{69.7}} \\
100\% & \textbf{\textcolor{red}{72.4}} & 70.9 & 69.4 & 70.8 \\
0\% & 69.1 & 50.1 & 49.6 & \textbf{\textcolor{red}{70.4}} \\
\bottomrule
\end{tabular}
}
\label{table:no}
\end{table*}

Misleading integrates multiple reasoning aspects from previous tasks and poses questions in an interrogative form. Each question specifies a claim about the image, such as the contraband category present, the number of instances, or the component configuration. These claims are intentionally designed to be plausible yet potentially false. The model must again choose \textit{Yes} or \textit{No} to judge the correctness of the statement. We intentionally ensure that questions with ``No'' as the correct answer constitute less than 50\% of the test set. 

\begin{figure*}[tbp]
\begin{promptbox}[Prompt of Option Redesign]
\footnotesize
\noindent\textbf{Instructions:}
You are an expert in lexical semantics and fine-grained category discrimination for contraband. Your task is to rank a list of candidate contraband categories by their textual semantic similarity to a given target contraband category, where the target serves as the correct answer. Higher ranked candidates should be semantically closer to the target in terms of functional purpose, typical carrying context, visual appearance, or security screening taxonomy. At the same time, each candidate must represent a distinct and discriminable contraband category. Synonymous labels and trivial variants that lack substantive distinction must be excluded from high ranking positions.
\medskip

\noindent\textbf{Example Input:}
\begin{lstlisting}[basicstyle=\ttfamily\small, breaklines=true, frame=single, numbers=none]
{
  "GT": "Glassbottleliquid",
  "Classes_list": ["Metalbottleliquid", "Plasticbottleliquid"...]
}
\end{lstlisting}

\noindent\textbf{Example Output:}
\begin{lstlisting}[basicstyle=\ttfamily\small, breaklines=true, frame=single, numbers=none]
{
  "GT": "Glassbottleliquid",
  "candidate_categories": ["Metalbottleliquid", "Compressedgas","Plasticbottleliquid"]
}
\end{lstlisting}

\medskip
\noindent\textbf{Now process the following input:}
\begin{lstlisting}[basicstyle=\ttfamily\small, breaklines=true, frame=none, numbers=none]
{
  "GT": "<TARGET>",
  "Classes_list": [<CLASS_1>, <CLASS_2>, ..., <CLASS_N>]
}
\end{lstlisting}
Return \textbf{only} the JSON output, no prefix, no commentary.
\end{promptbox}

\caption{Prompt of option redesign in instance identity.}
\label{fig:redesign}
\end{figure*}
Empirically, we observe a strong negative bias in most VLMs: when presented with X-ray contraband tasks without sufficient domain grounding, they tend to over-predict ``No''. 
This bias is confirmed by a experiment (Table~\ref{table:no}), when we restructure all questions to have ``No'' as the ground-truth answer, model accuracy artificially inflates, yet performance on our \textit{Basic Understanding} benchmark remains poor. 
This discrepancy validates that such gains are spurious, underscoring the necessity of our balanced design to prevent misleading evaluation.

\begin{table}[!tbp]
\centering
\footnotesize
\caption{The number of prompts for different VQA tasks on the evaluation benchmark of MMXray.}
\resizebox{0.8\linewidth}{!}{%
\begin{tabular}{l|c}
\toprule
Task & Number of Prompts \\
\midrule
Location & 3430\\
Counting & 2495\\
Identity & 2924 \\
Feature & 1545 \\ 
Misleading & 5848 \\
Basic understanding & 1427 \\ \hline
\rowcolor{gray!10}
All & 17669 \\
\bottomrule

\end{tabular}
}

\label{tab:task_dis}
\end{table}

Basic Understanding is a text only task that does not involve any image input. It assesses foundational domain knowledge about X-ray contraband detection. We constructed this subset by collecting publicly available technical materials on security screening and consulting 20 domain experts, each with over five years of frontline experience in contraband inspection. Using a vision language model as a knowledge organizer, we transformed this corpus into question answer pairs. The resulting questions include binary decisions (Yes or No) as well as single choice questions that require identifying the most accurate statement among several options.
The distribution of these six task types in the MMXray benchmark is summarized in Table~\ref{tab:task_dis}. In total we derive 17,669 question answer pairs from 2,924 images covering all six subtask types described above.

\begin{table*}[!tbp]
\centering
\small
\setlength{\tabcolsep}{4pt}
\caption{Ablation study on synthetic data usage across training stages. Performance is evaluated on the VQA task, with average accuracy as the overall metric. Subtask scores for \textbf{Instance Identity} and \textbf{Instance Feature} are emphasized due to their sensitivity to occlusion modeling. \textcolor{red}{Red} is the best result and \textcolor{blue}{Blue} is the second best result.}
\label{app:tab:dataab}
\resizebox{\textwidth}{!}{%
\begin{tabular}{l|ccccccc}
\toprule
Synthetic Data Strategy & Location & Counting & Identity & Feature & Misleading & Basic Understanding& Avg \\
\midrule
None in Stage 1 \& Stage 2 & 70.1 & 56.3 & 55.4 & 89.1 & 58.1 & 59.0 & 64.7  \\
Only in Stage 1 & 72.3 & 53.5 & 56.2 & 89.6 & 60.3 & \textbf{\textcolor{red}{60.1}} & 65.3 \\
Only in Stage 2 & \textbf{\textcolor{blue}{73.8}} & \textbf{\textcolor{blue}{60.1}} & \textbf{\textcolor{blue}{69.3}} & \textbf{\textcolor{blue}{90.7}} & \textbf{\textcolor{blue}{66.5}} & 60.0 & \textbf{\textcolor{blue}{70.1}} \\ \hline
\rowcolor{gray!10}
\textbf{Both Stage 1 \& Stage 2 (Ours)} & \textcolor{red}{\textbf{76.5}} &  \textcolor{red}{\textbf{60.7}} &  \textcolor{red}{\textbf{75.2}} &  \textcolor{red}{\textbf{94.8}} &  \textcolor{red}{\textbf{69.7}} & \textcolor{blue}{\textbf{60.0}} & \textcolor{red}{\textbf{72.8}}\\
\bottomrule
\end{tabular}
}
\end{table*}

\section{Ablation on Synthetic Data Usage}
\label{app:secsyn}
In the main text we introduce MMXray, which contains 3,000 images generated via AnyContraSyn. These synthetic samples are primarily designed to cover severe occlusion scenarios commonly encountered in real world security screening. To verify the reliability and necessity of this synthetic data we conduct a systematic data level ablation study as shown in Table~\ref{app:tab:dataab}. Specifically we compare model performance on the VQA task under four training data configurations:
(1) No synthetic data in either Stage 1 or Stage 2;
(2) Synthetic data included only in Stage 1;
(3) Synthetic data included only in Stage 2;
(4) Synthetic data included in both Stage 1 and Stage 2;

The results demonstrate that configuration (4) yields the best overall performance with notable gains on Instance Identity and Instance Feature. This improvement stems from the model's enhanced ability to learn component level features of contraband under color or contour distortion caused by occlusion. Our findings confirm the effectiveness and validity of the proposed synthetic data strategy.





\FloatBarrier

\begin{figure*}[p]
\begin{promptbox}[Prompt]
\scriptsize
\setlength{\parskip}{0pt}
\setlength{\baselineskip}{8.4pt}
\textbf{Background Description\\}
You are an authoritative expert in X-ray contraband detection and security screening with deep knowledge of:\\
- Aviation security regulations and contraband\\
- X-ray imaging principles and material discrimination (organic/inorganic/metal)\\
- Concealment tactics used by smugglers\\
- Standard operating procedures for security officers\\
- Common false positives and ambiguous threat signatures\\[0.5em]
Your task is to generate high-quality, multi-turn training conversations that will teach a vision-language model to excel at following four critical tasks: Visual Question Answering, Classification, Contraband Localization, and Image Understanding. You must create both open-ended and closed-ended questions that reflect real-world screening scenarios.\\
You have to generate all the conversations starting from the Image and Caption I gave at the very beginning.\\[0.5em]
\textbf{Task Description\\}
Given an X-ray security image and its caption, generate a multi-turn conversation (2-4 turns maximum) between a User (security officer) and Assistant (AI screening aid). Each conversation MUST contain:\\
1. At least ONE open-ended question with reasoning-rich answer\\
2. At least ONE closed-ended multiple-choice question\\
3. Content covering ALL six dimensions specified below\\
4. The output must include the Type comprising the Six Essential Dimensions.\\
The conversation must be realistic, pedagogically valuable, and adhere to strict output formats.\\[0.5em]
\textbf{The Six Essential Dimensions (Generate Content for Each)\\}
1. \textbf{Instance Location}: I will provide an X-ray scan image together with the corresponding contraband categories and bounding boxes information. Please generate QA pairs describing the location. Each answer should include spatial coordinates or area descriptors (e.g., “top-right quadrant”, “behind the laptop”, “center of the suitcase”, “left side in the image”).\\
2. \textbf{Instance Counting}: I will provide an X-ray scan image along with the corresponding categories and quantities of contraband items. Please generate QA pairs focusing on quantity calculation. Correct answers must account for occluded, overlapping, and stacked items. If a vague answer such as "some" is given, a justification must be provided.\\
3. \textbf{Instance Identity}: Given an X-ray scan image together with category information, the task is to generate a set of choice questions for category identification. Correct options must include all relevant categories and exclude incorrect ones. Each question should provide no more than five options.
The category of contrabands that may appear in the X-ray scan are as follows:
Mobilephone, 
Powerbank, 
Handcuffs,
Axe,
Scissor, 
Baton, 
Bullet, 
Hammer, 
Gun, 
Metalhandlecleaver, 
Plastichandlecleaver, 
Pliers,
Compressdgas, 
Columnarblockbattery, 
Plasticbottleliquid,
Glassbottleliquid,
Metalbottleliquid, 
UnfoldingKnife, 
FoldingKnife, 
StraightKnife, 
UtilityKnife, 
MultitoolKnife, 
PlastichandleKnife, 
MetalhandleKnife, 
Platiclighter, 
Laptop, 
Grenade, 
Fireworks.
\\
4. \textbf{Instance Feature}: I will provide an X-ray scan image containing contraband items together with physical attributes critical for threat assessment,  such as dimension, shape, material composition (metal/organic/inorganic), and density indication.    Generate single-choice QA conversations focusing on these attributes of the contrabands in the image to test whether the model can recognize such features.\\
5. \textbf{Misleading Context}: I will provide an X-ray scan image together with the corresponding category label.    Please generate highly misleading QA pairs by introducing nonexistent categories or objects into the questions to test the model's ability to handle deliberately deceptive arrangements.
Or generate QA pairs related to deceptive items or attributes,for example,items masked by other objects, innocent-item appearance (e.g., “gun-shaped lighter”),anomalies in material presentation, potential false positives, and contraband mimicking everyday objects.\\
6. \textbf{Basic Understanding}: Different contraband categories exhibit distinct basic attributes such as shape and color.   Generate single-choice QA pairs to test the basic knowledge of contraband without reference to images.   These QA pairs may either specify attributes in the question with categories as options, or specify categories in the question with features as options.\\[0.5em]
\textbf{Open-Ended Question Requirements\\}
- Questions must be natural and varied, not templated\\
- Answers MUST include explicit reasoning chain: Observation $\rightarrow$ Analysis $\rightarrow$ Conclusion\\
- Minimum 2--3 sentences per answer with technical detail\\
- Use security domain terminology accurately\\
- For location tasks, provide coordinate-like precision\\
- For ambiguous cases, explain uncertainty factors\\[0.5em]
\textbf{Closed-Ended Question Requirements\\}
- Provide EXACTLY 2--4 answer choices (labeled A, B, C, D)\\
- ONLY ONE correct answer --- no ambiguity\\
- Choices must be plausible distractors based on common misconceptions\\
- Assistant's response must be ONLY the letter of the correct choice (e.g., ``C'')\\
- Include a brief explanation AFTER the letter in parentheses: ``(C) Because the metallic signature and trigger guard shape are definitive indicators''\\[0.5em]
\textbf{Output Format(Strict JSON Structure)\\}
\{\\
\ \ "conversation": [\\
\ \ \ \ \{"role": "User", "content": "question1"\},\\
\ \ \ \ \{"role": "Assistant", "content": "answer1"\},\\
\ \ \ \ \{"role": "User", "content": "question2"\},\\
\ \ \ \ \{"role": "Assistant", "content": "answer2"\},\\
\ \ \ \ // 2--4 turns total\\
\ \ ],\\
\ \ "image": [Path],\\
\ \ "Type": [Six Essential Dimensions]\\
\}
\end{promptbox}
\vspace{-0.6em}
\caption{VLM training prompt for X-ray contraband detection (used to generate multi-turn conversations).}
\label{fig:prompt}
\end{figure*}

\end{document}